%% file: paper.tex
\documentclass[runningheads]{llncs}
\usepackage[T1]{fontenc}
\usepackage{graphicx}
\usepackage{amssymb}
\usepackage{bm}
\usepackage{algorithmic}
\usepackage{algorithm}
\input{macro}

\usepackage{wrapfig}

\begin{document}
\title{Tunable MAGMAX: Preference-Aware Model Merging for Continual Learning}
\titlerunning{Preference-Aware Model Merging for Continual Learning}
\author{Kei Hiroshima\and
Kento Uchida\and  
Shinichi Shirakawa} 
\authorrunning{K. Hiroshima et al.}
%
\institute{Yokohama National University, Yokohama Kanagawa 240-8501, Japan}
%
\maketitle              
\begin{abstract}

Continual learning (CL) aims to train models sequentially on multiple tasks while mitigating catastrophic forgetting of previously learned knowledge.
Recent advances in large pre-trained models (LPMs) and model merging techniques, such as MAGMAX, have demonstrated effective CL performance by combining task-specific parameters.
However, existing methods primarily focus on average performance across all tasks and do not adequately address how to construct models accommodating different deployment environments or varying user preferences.
This paper proposes a model merging framework, termed Tunable MAGMAX, which enables preference-aware control of task-specific performance in CL.
Our method introduces a preference vector that controls the number of elements selected from each task vector during model merging, allowing us to adjust the merged model performance according to their deployment needs.
We further propose a method for automatically constructing appropriate preference vectors by leveraging small amounts of target environment data and datasets from model training tasks, thereby eliminating the need for manual specification.
The experimental result on CL benchmark tasks demonstrates that Tunable MAGMAX effectively controls task-wise performance and successfully adapts merged models to various target environments.
The proposed Tunable MAGMAX achieves superior or comparable performance to baseline methods, making it a practical solution for deploying CL models to various environments where the preferences of each task performance differ.

\end{abstract}

\section{Introduction}\label{sec:intro}

In continual learning (CL), a machine learning model is trained sequentially on a series of tasks without access to data from previous tasks~\cite{cl_survey}. The objective of CL is to obtain a generalized model across all tasks by adapting to incoming tasks while retaining knowledge from previously trained tasks. Models trained in CL often suffer from catastrophic forgetting (CF), a phenomenon where performance on previous tasks degrades due to biased training on the current task.
To alleviate CF, many CL techniques have been developed. The major techniques can be categorized into three groups: adding regularization terms to the loss function to maintain knowledge of previously trained tasks~\cite{ewc,lwf}, expanding the model architecture to avoid parameter interference among training tasks~\cite{pnn}, and utilizing replay buffers that store small amounts of data from previous tasks to enable retraining~\cite{mir}. 

These traditional techniques have mainly been evaluated on relatively simple models with limited capacity, such as deep and convolutional neural networks.
Nowadays, large pre-trained models (LPMs), including large-scale Vision Transformers pre-trained on extensive datasets (e.g., ImageNet~\cite{imagenet} and LAION-400M~\cite{laion400m}), have become a common tool, owing to rapid advancements in deep learning technology~\cite{Caron_2021_ICCV,corr}. 
These models have a large capacity and perform remarkably well on a wide variety of tasks.

In addition, model merging, a cost-efficient approach to developing models tailored to specific tasks~\cite{akiba_evolutionary_2024,bandarkar2025layer}, has been investigated~\cite{task_arithmetic,ties_merging}.
Model merging creates a new model tailored to a given task by combining multiple model parameters fine-tuned on different tasks from a common LPM.

LPMs are also adopted as base models in CL. In such scenarios, it is crucial not only to mitigate CF during training but also to effectively leverage the general capabilities of LPMs. Maximum Magnitude Selection (MAGMAX)~\cite{marczak_magmax_2025} demonstrated the effectiveness of LPMs and model merging in CL by utilizing model parameters corresponding to sequentially fine-tuned models across a sequence of tasks. 
In MAGMAX, the task vector is defined as the difference between the base pre-trained model parameters and those of the sequentially fine-tuned model from the base model. The task vectors obtained for each trained task are merged by selecting the value with element-wise maximum absolute magnitude across the task vectors. 
Despite having relatively low space and time complexities, MAGMAX outperforms other CL and model merging methods.

Previous works on CL have focused on improving average performance across all trained tasks. In contrast, the importance of each task can vary depending on user demands or deployment environments in practical situations. 
For instance, in medical diagnosis systems, different hospitals may prioritize different disease detection tasks based on their specialization, such as tumor detection or cardiac abnormality detection. However, due to limited computational resources and insufficient training data available at each hospital, they often rely on a published pre-trained model trained on general medical tasks and customize it to emphasize only the most relevant tasks for their specific needs.
Similarly, in autonomous driving, vehicles deployed in different geographical regions, such as rainy or snowy climates, or urban versus highway environments, require different emphases on region-specific driving tasks, even though the base LPM is pre-trained to cover a wide range of driving conditions during pre-training.
From this perspective, existing CL techniques have not adequately addressed how to construct models flexibly to reflect these diverse and deployment-specific task priorities.
In that sense, it is unclear whether MAGMAX~\cite{marczak_magmax_2025} remains effective. 
In fact, Marczak et al.~\cite{marczak_magmax_2025} reported that the task vector corresponding to the last trained task accounted for more than 40\% of the elements of the merged task vector.
This dominance can lead to inferior performance compared to other existing methods on some tasks, even though MAGMAX achieves the highest average performance.

In this paper, we propose a novel model-merging-based CL method for preference-aware control of performances for each task by extending MAGMAX.
The proposed method, Tunable MAGMAX, adjusts the number of elements selected from each task vector during model merging based on a preference vector.
We further introduce a method that automatically sets the preference vector and constructs an appropriate merged model tailored to various target environments with a lower computational cost than training their own model from scratch.
In our method, a preference vector is determined based on the dataset similarity, such as one defined using the optimal transport (OT) distance~\cite{Kantorovich2006OnTT} and the label distribution similarity, between a small amount of target environment data and datasets from the model training tasks.
We conduct validation experiments based on CL benchmarks to compare our proposed method with the existing model merging methods. 
Experimental results demonstrate that our proposed method successfully controls task-wise performance, and our method constructs suitable merged models for given target environments by controlling the preference vector based on dataset similarity.

Our contributions are summarized as follows:

\begin{itemize}
    \item We propose Tunable MAGMAX, a preference-aware model merging for CL, that regulates the number of elements selected from each task vector to control task-wise performance.
    \item We propose a method to construct merged models tailored to target environments based on dataset similarity between small amounts of target environment data and datasets from model training tasks.
    \item Through the experimental evaluation on CL benchmarks, we demonstrate that our framework successfully constructs preferable merged models for various target environments. 
    
\end{itemize}

\section{Related Works}
\subsection{Continual Learning}

In continual learning (CL) settings, a single model is repeatedly trained on multiple tasks in a sequence while being required to maintain its knowledge from previously trained tasks without access to training data from those tasks. The domain, class, and task incremental learning are widely known CL settings in classification tasks~\cite{cl_survey}. In domain incremental learning, tasks do not share their input distribution but share the same label space. In class incremental learning, tasks involve different label spaces while sharing the same input domain. In the task incremental learning, tasks share neither the input domain nor the label space.

Previous approaches to mitigating the problem of catastrophic forgetting (CF) in CL settings are categorized into three groups based on their methodologies: regularization-based approaches, replay-based approaches, and parameter isolation-based approaches. Elastic Weight Consolidation (EWC)~\cite{ewc}, which is a regularization-based approach, prevents certain model parameters from being updated during training on later tasks. These parameters are selected based on their importance defined by an approximated Fisher information matrix calculated at the end of training on each task. Learning without Forgetting (LwF)~\cite{lwf} is another regularization-based approach that penalizes model updates to prevent drastic changes in the model's outputs during training on later tasks. Maximally Interfered Retrieval (MIR)~\cite{mir}, which is a replay-based approach, alleviates CF by training on new task data combined with a small amount of valuable data from previously trained tasks stored in a replay buffer. Progressive Neural Networks~\cite{pnn}, a parameter isolation approach, expand the model architecture by allocating new parameters for training on new tasks, thereby preserving model parameters obtained during training on previous tasks.

\subsection{MAGMAX} \label{sec:magmax}

Maximum Magnitude Selection (MAGMAX)~\cite{marczak_magmax_2025} is a CL method based on model merging of fine-tuned LPMs. In MAGMAX, the LPM is sequentially fine-tuned on a sequence of tasks~$\{\mathcal{D}_1, \ldots, \mathcal{D}_T\}$, and the model parameters for each task~$\bm{\theta}_1, \ldots, \bm{\theta}_T \in \mathbb{R}^d$ are obtained. Then the task vectors for these tasks are computed, which are defined as $\bm{\tau}_k = \bm{\theta}_k - \bm{\theta}_0 \quad \text{for} \quad k = 1,\ldots,T$, where $\bm{\theta}_0 \in \mathbb{R}^d$ represents the LPM's parameters before fine-tuning.

In the model merging phase, task vectors are merged by selecting the value with the maximum absolute magnitude in the element-wise manner, i.e., for $p=1,\ldots,d$, the element of the merged task vector is computed as
\begin{align}
    [\bm{\tau}_{\mathrm{merged}} ]_p  = \sum_{k=1}^T [\bm{\tau}_k]_p \cdot \mathbbm{1}{\{ k = \argmax_{i=1,\ldots,T} \{ | [\bm{\tau}_i]_p | \} \}} \enspace,
\end{align}
where $[\bm{a}]_p$ denotes the $p$-th element of a given vector $\bm{a}$, and $\mathbbm{1}\{E\} \in \{0,1\}$ is the indicator function for an event $E$.
In addition, by iteratively merging the task vector $\bm{\tau}_k$ for the current task and the merged task vector $\bm{\tau}_{\mathrm{merged}, k-1}$ so far, MAGMAX can be performed with constant memory with respect to the number of tasks.

According to experiments in~\cite{marczak_magmax_2025}, MAGMAX was demonstrated to perform better than other CL and model merging approaches. 
However, MAGMAX does not consider the importance of trained tasks, which can vary depending on user demands or deployment environments in practical situations.
Therefore, it is necessary to extend MAGMAX for preference-aware model construction to reflect user demands.

\begin{figure}[t!]
    \centering
    \includegraphics[width=0.95\linewidth]{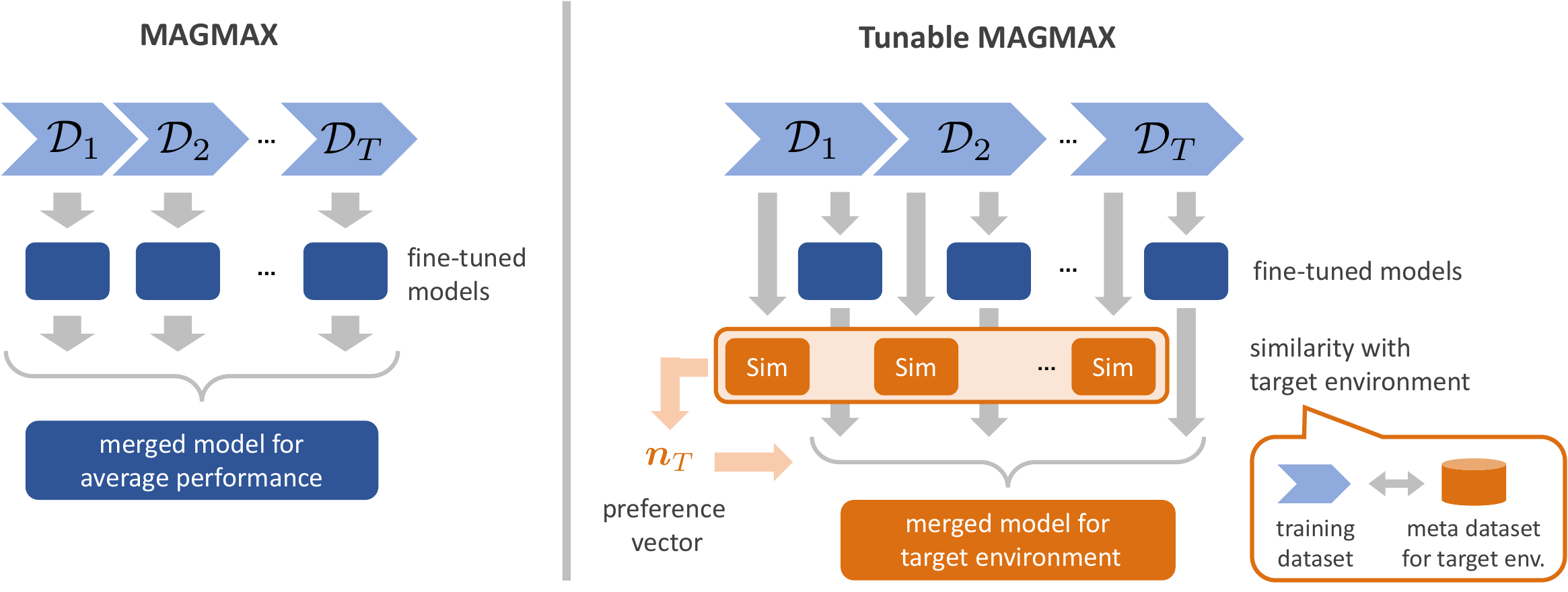} 
    \caption{Our method constructs the merged model tailored to the target environment according to the preference vector $\bm{n}_T$ generated by dataset similarity, while MAGMAX~\cite{marczak_magmax_2025} constructs the merged model that works well on all training tasks on average.}
    \label{fig:setting}
\end{figure} 

\section{Proposed Method}
We explain our problem setting in Section~\ref{sec:problem_setting}, our proposed method called Tunable MAGMAX that controls task-wise performance via preference vector during model merging in  Section~\ref{sec:proposed:num_elememt}, and how to compute the preference vector in Section~\ref{sec:proposed:main}.
Figure~\ref{fig:setting} illustrates our approach.

\subsection{Problem Setting} \label{sec:problem_setting}

We consider a traditional CL scenario where a model sequentially learns from a stream of disjoint tasks $\{\mathcal{D}_1, \mathcal{D}_2, \ldots, \mathcal{D}_T\}$, where $T$ denotes the total number of tasks.
Each task has pairs of features and ground-truth labels: $\mathcal{D}_t=\{(x_t^{(j)},y_t^{(j)})\}_{j=1}^{N_t}$, where $x_t^{(j)} \in \mathcal{X}_t$ and $y_t^{(j)} \in \mathcal{Y}_t$ with feature space $\mathcal{X}_t$ and label space $\mathcal{Y}_t$ for $\mathcal{D}_t$, and $N_t$ is the number of data in the $t$-th task $\mathcal{D}_t$. 
For deployment, we consider the target dataset on the target environment $\mathcal{D}_\text{tar} = \{ (x_\text{tar}^{(j)},y_\text{tar}^{(j)}) \}_{j=1}^{N_\text{tar}}$, where $N_\text{tar}$ is the number of data in $\mathcal{D}_{\mathrm{tar}}$ which can be varied. The target environment $\mathcal{D}_\text{tar}$ has its own data distribution that may not be identical to any trained tasks $\{\mathcal{D}_t \}^T_{t=1}$.
Our aim is to construct an appropriate model for each target environment without retraining the fine-tuned model in CL.
Additionally, we consider the following scenarios: 1) the meta dataset $\mathcal{D}_\text{meta} = \{ (x_\text{meta}^{(j)},y_\text{meta}^{(j)}) \}_{j=1}^{N_\text{meta}}$ on the target environment is accessible, 2) the feature set $\mathcal{D}_{\text{meta}, x} = \{ x_\text{meta}^{(j)} \}_{j=1}^{N_\text{meta}}$ is accessible, 3) the label set $\mathcal{D}_{\text{meta}, y} = \{ y_\text{meta}^{(j)} \}_{j=1}^{N_\text{meta}}$ is accessible, and 4) no data is accessible, where $N_\text{meta}$ is the number of data in $\mathcal{D}_\text{meta}$.

We assume access to a pre-trained model with parameters $\bm{\theta}_0 \in \mathbb{R}^d$, where $d$ represents the total number of parameters in the model. 
For consecutive tasks $t \in \{1, 2, \ldots, T\}$, we sequentially fine-tune the model using each task-specific training dataset to obtain task-specific parameters $\bm{\theta}_t$. Following the task vector framework~\cite{task_arithmetic}, we define the task vector for task $t$ as the element-wise difference between the fine-tuned and pre-trained model parameters as $\bm{\tau}_t = \bm{\theta}_t - \bm{\theta}_0 \in \mathbb{R}^d$. The task vector $\bm{\tau}_t$ encapsulates the knowledge specific to the $t$-th task.

\begin{algorithm}[t!]
	\caption{Merge Process of Tunable MAGMAX}
	\label{alg:merge}
    \begin{algorithmic}[1]
    \REQUIRE Task vectors $\bm{\tau}_1, \cdots, \bm{\tau}_T$, preference vector $\bm{n}_T = (n_1,\ldots,n_T)$, number of assignment processes $K=2$
    \STATE Set $\mathcal{I}_\mathrm{unselect} = \{ 1, \cdots, d \}$ and $\mathcal{S}_t = \emptyset$ for all $t=1,\ldots,T$
    \FOR{$k=1,\ldots,K$}
        \FOR{$t = T,\ldots, 1$}
            \STATE $\hat{\mathcal{S}}_t \leftarrow$ Elements in $\mathcal{I}_\mathrm{unselect}$ where $\bm{\tau}_t$ has largest magnitude among $\{\bm{\tau}_{t'}\}_{t'=1}^t$.
            
            \IF{ $|\mathcal{S}_t \cup \hat{\mathcal{S}}_t| > n_t$ }
                \STATE $\hat{\mathcal{S}}_t \leftarrow$ Select $n_t - |\mathcal{S}_t| $ elements from $\hat{\mathcal{S}}_t$ randomly
            \ENDIF
            \STATE $\mathcal{S}_t \leftarrow \mathcal{S}_t \cup \hat{\mathcal{S}}_t$ and $\mathcal{I}_\mathrm{unselect} \leftarrow \mathcal{I}_\mathrm{unselect} \setminus \hat{\mathcal{S}}_t$
        \ENDFOR
    \ENDFOR
    \FOR{$t = 1,\ldots, T$}
        \STATE $\hat{\mathcal{S}}_t \leftarrow$ Select $n_t - |\mathcal{S}_t| $ elements from $\mathcal{I}_\mathrm{unselect}$ randomly
        \STATE $\mathcal{S}_t \leftarrow \mathcal{S}_t \cup \hat{\mathcal{S}}_t$ and $\mathcal{I}_\mathrm{unselect} \leftarrow \mathcal{I}_\mathrm{unselect} \setminus \hat{\mathcal{S}}_t$
    \ENDFOR
    \FOR{$t = 1,\ldots, T$}
        \STATE $[\bm{\tau}_{\mathrm{merged}} ]_{p} = [\bm{\tau}_{t}]_{p}$ for all $p \in \mathcal{S}_t$
    \ENDFOR
    \RETURN Merged task vector $\bm{\tau}_{\mathrm{merged}}$
\end{algorithmic}
\end{algorithm}

\subsection{Tunable MAGMAX} \label{sec:proposed:num_elememt}
In the model merging method proposed in MAGMAX~\cite{marczak_magmax_2025}, the number of elements selected from each task vector for the merged task vector is not explicitly controlled, which can result in bias towards certain task vectors. To address this issue, we propose a new model merging method called Tunable MAGMAX that utilizes a preference vector $\bm{n}_T = (n_1, \ldots, n_T) \in \mathbb{N}^T$.
The preference vector is set to satisfy $\sum^{T}_{t=1} n_{t} = d$, where $d$ denotes the number of model parameters, and is used to adjust the number of elements selected from $T$ task vectors $\{\bm{\tau}_t\}_{t=1}^T$ for the merged task vector during model merging. We can regard the $t$-th value in the preference vector as the importance of the $t$-th task in the merged model.

In the merging process, Tunable MAGMAX selects the elements corresponding to the $t$-th task vector $\bm{\tau}_t$ in descending order $t = T, \cdots, 1$ as follows.
First, for the $t$-th selection phase, the elements that are not selected before and have the largest magnitude in the $t$-th task vector among $\bm{\tau}_1,\cdots,\bm{\tau}_t$ are selected as
\begin{align}
\label{eq:taskvec_sel}
    \hat{\mathcal{S}}_t = \left\{ p \in \mathcal{I} \setminus \mathcal{S}_{t+1:T} \, \middle| \, t = \argmax_{t'=1,\ldots,t} \{ | [\bm{\tau}_{t'}]_{p} | \} \right\} \enspace,
\end{align}
where $\mathcal{I} = \{1, \ldots, d\} $ and $\mathcal{S}_{t+1:T} = \bigcup_{t'=t+1,\ldots,T} \mathcal{S}_{t'}$ are the indexes of all elements and the elements selected before the $t$-th task vector, respectively.
If the number of selected elements $|\hat{\mathcal{S}}_t|$ is greater than $n_t$, $n_t$ elements are selected randomly as
\begin{align}
\label{eq:red_elem}
    \mathcal{S}_t = \begin{cases}
    \hat{\mathcal{S}}_t & \text{if $|\hat{\mathcal{S}}_t| \leq n_t$} \\
    \text{$n_t$ elements in $\hat{\mathcal{S}}_t$ randomly selected} & \text{otherwise} \enspace.
    \end{cases}
\end{align}
Then, the elements of the merged task vector are given by the selected elements in $\mathcal{S}_t$ as
\begin{align} 
    [\bm{\tau}_{\mathrm{merged}} ]_{p} = [\bm{\tau}_{t}]_{p} \quad \text{for} \quad p \in \mathcal{S}_t \enspace \text{and} \enspace t=1 \cdots, T \enspace.
\end{align}

This adjustment is performed starting from the last task because the task vector $\bm{\tau}_T$ corresponding to the last trained task was reported to account for a large part of the merged task vector in MAGMAX~\cite{marczak_magmax_2025}. 
We note that, when there is at least one task vector $\bm{\tau}_t$ for which the number of selected elements $|\mathcal{S}_t|$ is smaller than its corresponding budget $n_{t}$ in the preference vector, some elements may remain unassigned after the assignment process by \eqref{eq:taskvec_sel} and \eqref{eq:red_elem}.
To resolve this issue, we repeat the assignment process several times to assign the unselected elements to task vectors whose selected elements are fewer than the value of the preference vector as $|\mathcal{S}_k | < n_k$.
After repeating the assignment for $K$ times, the remaining elements are randomly assigned to task vectors whose selected elements are still smaller than the preference vector $| \mathcal{S}_k | < n_k$. We set $K=2$ in this paper. 
We empirically confirmed that randomly assigned parameters were zero in many cases, or at most 10\% of the whole model, and that these parameters did not significantly affect merged models' performance. Therefore, this randomness does not critically impact overall model performance.

Algorithm~\ref{alg:merge} shows the merge procedure of Tunable MAGMAX.
With this algorithm, the merged task vector can involve important elements from each task vector while keeping the number of elements selected from the task vector consistent with the preference vector. As a result, Tunable MAGMAX enables us to control the merged model's task-wise performance flexibly with the preference vector.

\subsection{Similarity-Based Preference Control}
\label{sec:proposed:main}
The choice of the preference vector $\bm{n}_T$ affects the task-wise performance of the merged model produced by Tunable MAGMAX.
It is possible to manually construct an appropriate preference vector for the target environment. However, it may be tedious work.
Therefore, we propose two mechanisms for constructing the preference vector using a small amount of meta data $\mathcal{D}_{\mathrm{meta}}$ collected from a target environment.
Our mechanisms compute the preference vector using the dataset similarities between the meta dataset and training datasets for each task.
Our mechanisms focus on the case where either the feature set $\mathcal{D}_{\text{meta}, x} = \{ x_\text{meta}^{(j)} \}_{j=1}^{N_\text{meta}}$ or the label set $\mathcal{D}_{\text{meta}, y} = \{ y_\text{meta}^{(j)} \}_{j=1}^{N_\text{meta}}$ is accessible. 
We note that, when pairs of features and labels $\mathcal{D}_\text{meta} = \{ (x_\text{meta}^{(j)},y_\text{meta}^{(j)}) \}_{j=1}^{N_\text{meta}}$ are given, the preference vector can alternatively be tuned via hyperparameter optimization.

\input{pseudo_whole}

\subsubsection{Preference Control with Feature Set} 
When only the feature set in $\mathcal{D}_{\mathrm{meta},x}$ is available, we use the optimal transport (OT) distance~\cite{Kantorovich2006OnTT} of the embedded features as the dataset similarity.
Specifically, the similarity $\mathrm{sim}(\mathcal{D}_t,\mathcal{D}_{\mathrm{meta},x})$ between the $t$-th task $\mathcal{D}_t$ and the meta dataset $\mathcal{D}_{\mathrm{meta},x}$ is defined as
\begin{align*}
\mathrm{sim}(\mathcal{D}_t,\mathcal{D}_{\mathrm{meta},x}) = \exp \left( - \gamma \cdot \mathrm{OT} \left( \left\{ \bm{f}_{\bm{\theta}_t}(x_{t}^{(j)}) \right\} {}_{j=1}^{N_t},~ \left\{\bm{f}_{\bm{\theta}_t}(x_{\mathrm{meta}}^{(j)}) \right\} {}_{j=1}^{N_{\mathrm{meta}}} \right) \right) \enspace,
\end{align*}
where $\bm{f}_{\bm{\theta}_t}$ denotes the image encoder with the parameter $\bm{\theta}_t$ and $\gamma$ is a scaling parameter. We set the scaling parameter as $\gamma=100$ in this paper. The OT distance $\mathrm{OT}(\cdot,\cdot)$ is approximated using the Sinkhorn algorithm~\cite{cuturi_sinkhorn_2013}.
The preference vector $\bm{n}_T$ is then constructed by allocating elements proportionally to the computed similarities as
\begin{align} \label{eq:sim_to_pv}
    n_t &= \bar{n}_t+ \mathbbm{1}\{t \leq R \} \enspace,
\end{align}
where
\begin{align} \label{eq:remainder}
    \bar{n}_t = \left\lfloor \frac{\mathrm{sim}(\mathcal{D}_t,\mathcal{D}_{\mathrm{meta},x})}{\sum_{t'=1}^T \mathrm{sim}(\mathcal{D}_{t'},\mathcal{D}_{\mathrm{meta},x})} \cdot d \right\rfloor 
    \quad \text{and} \quad 
    R = d - \sum_{t=1}^T \bar{n}_t \enspace.
\end{align}
We note that $\sum_{t=1}^T n_t = d$ is always satisfied.

\subsubsection{Preference Control with Label Set}
When only the label set $\mathcal{D}_{\mathrm{meta},y}$ is available, we use the label distribution similarity between each training task and the target environment, which is defined as
\begin{align}\label{eq:dist_to_scv}
    \mathrm{sim}(\mathcal{D}_t,\mathcal{D}_{\mathrm{meta},y}) = \sum_{c \in \mathcal{Y}_\mathrm{meta}} r_c(\mathcal{D}_t) \cdot r_c(\mathcal{D}_{\mathrm{meta},y}) \enspace,
\end{align}
where $\mathcal{Y}_\mathrm{meta}$ is the set of classes that appear in the meta dataset, and $r_c(\mathcal{D}) = \sum_{j=1}^{|\mathcal{D}|} \mathbb{I}\{ y^{(j)} = c \} / |\mathcal{D}|$ is the empirical appearance probability of class $c$ in the dataset $\mathcal{D}$.
We note that this similarity corresponds to the inner product between empirical label distributions.
The preference vector is computed by \eqref{eq:sim_to_pv} using the label-distribution similarities.

\subsubsection{Tunable MAGMAX with Similarity-Based Preference Control}
Algorithm~\ref{alg:whole} provides the overall procedure of Tunable MAGMAX with dataset similarity-based construction of the preference vector. 
With this procedure, our method enables users to construct merged models without requiring manual specification of preference vectors for various target environments, while maintaining reasonable computational overhead.

\section{Experiments and Results}
In this section, we evaluate the proposed Tunable MAGMAX on standard CL benchmarks.
Section~\ref{sec:exp_setting} explains the experimental setting. Section~\ref{sec:res:vs_magmax} presents the experimental results of Tunable MAGMAX with exemplary preference vectors.
Section~\ref{sec:main_results} evaluates the similarity-based preference control mechanism.
Finally, Section~\ref{sec:ablation} shows the result of the ablation study.

\begin{table}[t!]
\centering
\caption{Experimental setting for CIFAR-100 and ImageNet-R.}
\label{tab:setting}
\begin{tabular}{l @{\hspace{3pt}}|@{\hspace{3pt}} lll@{\hspace{3pt}}|@{\hspace{3pt}}lll}
\toprule
& \multicolumn{3}{c}{CIFAR-100-$T$} & \multicolumn{3}{c}{ImageNet-R-$T$}\\
\midrule
Number of tasks ($T$) & 5 & 20 & 50 & 5 & 20 & 50 \\
Number of classes per task & 20 & 5 & 2 & 40 & 10 & 4 \\
Number of training data per task & 10,000 & 2,500 & 1,000 & 4,800 & 3,000 & 480 \\
Number of test data per task & 2,000 & 500 & 200 & 1,200 & 600 & 120 \\
\bottomrule
\end{tabular}
\end{table}

\subsection{Experimental Setting}\label{sec:exp_setting}
\subsubsection{Datasets}

Following~\cite{marczak_magmax_2025}, we conduct experiments in class incremental learning settings.
We use CIFAR-100~\cite{cifar100} and ImageNet-R~\cite{hendrycks2021many}, which have 100 and 200 classes, respectively.
For each dataset, we divide it into $T$ equal subsets of disjoint classes, where $T \in \{5, 20, 50\}$. We denote each task sequence setting as CIFAR-100-$T$ and ImageNet-R-$T$. 
Table~\ref{tab:setting} shows the detailed settings for each dataset.
The order of tasks in the task sequence is set as well as~\cite{marczak_magmax_2025}. 

\subsubsection{Implementation Details}
We use the CLIP~\cite{clip} with a ViT-B/16~\cite{vit} image encoder as a pre-trained model.
We fine-tune the image encoder using a batch size of 128 and the AdamW optimizer with a learning rate of $10^{-5}$ and a cosine annealing scheduler. Models train on each task for 10 epochs. The final classification layer is imported from CLIP's text encoder and kept frozen during training so that the model can preserve its vocabulary obtained during pre-training. These experimental settings follow~\cite{marczak_magmax_2025}, and the implementation is based on its official code\footnote{\url{https://github.com/danielm1405/magmax/}}. The code of Tunable MAGMAX will be made available at \url{https://github.com/KeiHiroshima/tunable-magmax}.

\begin{figure}[t!]
    \centering
    \includegraphics[width=0.55\textwidth]{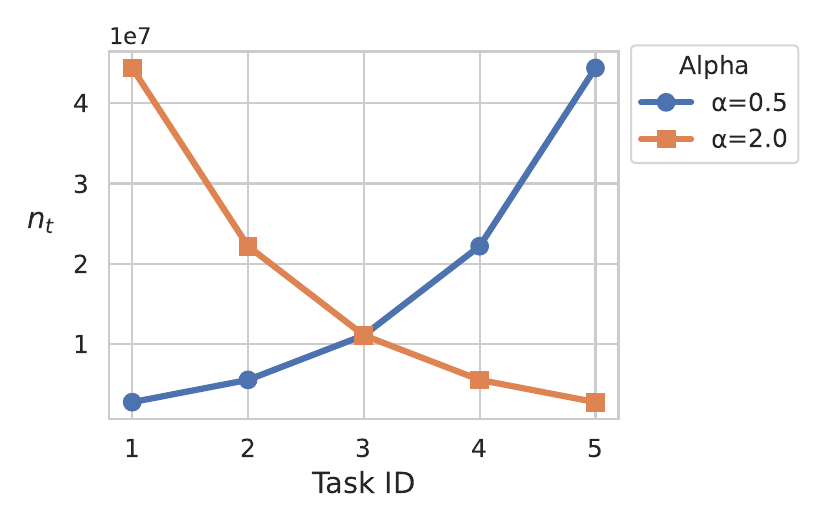}
    \caption{Preference vector for CIFAR-100-5 with $\alpha=0.5, 2.0$ (and $d \approx 86 \times 10^6$).}
    \label{fig:num_elements_alpha}
\end{figure}

\subsection{Evaluation of Tunable MAGMAX with Exemplary Preference Vectors}\label{sec:res:vs_magmax}
In this section, we evaluate the flexibility of the task-wise performance of the merged model constructed by Tunable MAGMAX.
As an exemplary setting of the preference vector, we introduce a coefficient $\alpha \geq 0$ and define the $t$-th element of the preference vector $\bm{n}_T$ as
\begin{align}
    \label{eq:n_with_alpha}
    & n_t = \bar{n}_t + \mathbb{I}\{ t \leq R \} \\
    \label{eq:n_with_alpha2}
    & \text{where} \quad \bar{n}_t = \left\lfloor\frac{\alpha^{T-t}}{\sum^{T}_{t'=1}\alpha^{T-t'}} \cdot d \right\rfloor  \quad \text{and} \quad R = d - \sum_{t=1}^T \bar{n}_t \enspace.
\end{align}
Figure~\ref{fig:num_elements_alpha} illustrates the resulting preference vector for CIFAR-100-5 with $\alpha=0.5$ and $\alpha=2$.
When $\alpha > 1$, the task vectors corresponding to earlier tasks contribute more elements to the merged model, whereas $\alpha < 1$ prioritizes later tasks.
We set $\bm{n}_T = (0, \ldots, 0, d)$ when $\alpha=0$.
The case $\alpha = 1$ results in an equal contribution from all task vectors.

\begin{figure}[t!]
    \centering
    \includegraphics[width=0.995\linewidth]{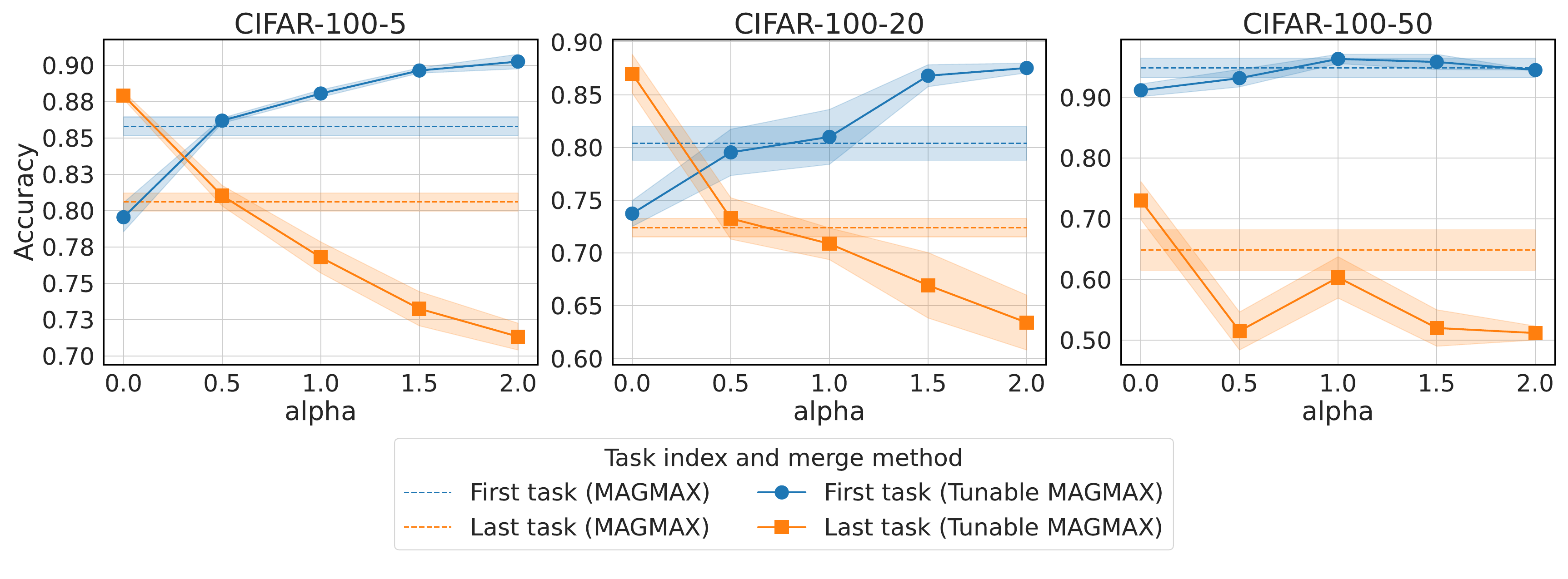}
    \caption{Top-1 accuracy for the first and last tasks on CIFAR-100-5 (left), -20 (center), -50 (right) with the preference vectors in \eqref{eq:n_with_alpha} and \eqref{eq:n_with_alpha2} parametrized by $\alpha$. The mean and standard deviation over three runs are plotted for each setting.}
    \label{fig:alpha_cifar100}
\end{figure}

We compare our proposed method (\textbf{Tunable MAGMAX}) with \textbf{MAGMAX}~\cite{marczak_magmax_2025} in terms of task-wise performance to demonstrate the controllability of the preference vector.
We set the scaling factor $\lambda_{\mathrm{merge}}=0.5$ for both methods, and a merged model $\bm{\theta}_{\mathrm{merged}}$ is given as follow: $\bm{\theta}_{\mathrm{merged}} = \bm{\theta}_0 + \lambda_{\mathrm{merge}}\cdot \bm{\tau}_{\mathrm{merged}}$.
All other experimental settings follow those described in Section~\ref{sec:exp_setting}.
Figure~\ref{fig:alpha_cifar100} presents the average top-1 accuracy on test data from the first and last tasks on CIFAR-100-$T$ with $T \in \{5, 20, 50\}$ among three different random seeds.
When $\alpha =2.0$, our proposed method performs better than MAGMAX on the first task, whereas when $\alpha =0.5$, our proposed method performs better or comparable compared to MAGMAX on the last task of each setting. 
These results demonstrate that the control of the number of elements selected from each task vector affects the model performance on the corresponding task, and Tunable MAGMAX flexibly handles the performance trade-off between earlier and later tasks as the preference vector varies.

\subsection{Evaluation of Tunable MAGMAX with Similarity-based Preference Control}\label{sec:main_results}
To evaluate the effectiveness of our similarity-based preference control mechanisms, we construct multiple target environments for CIFAR-100-$T$ and ImageNet-R-$T$ that simulate various real-world conditions where models encounter different data distributions from their training tasks.
Following the problem setting described in Section~\ref{sec:problem_setting}, we design target environments with five different variations.
These target environments are constructed by mixing test datasets for $M \in \{2, 3\}$ tasks with a pre-defined mixing ratio $\bm{a}_{\text{mix}} = (a_1, \ldots, a_M)$ that satisfies $\sum_{j=1}^{M} a_j = 1$.
We prepare five configurations of the target environments $\mathcal{D}_{\mathrm{tar},i}$ ($i \in \{1, \ldots, 5\}$) as follows.
\begin{enumerate}
    \item $\mathcal{D}_{\mathrm{tar},1}$: balanced mixture of two tasks ($\bm{a}_{\text{mix}} = (0.5, 0.5)$)
    \item $\mathcal{D}_{\mathrm{tar},2}$: imbalanced mixture of two tasks ($\bm{a}_{\text{mix}} = (0.8, 0.2)$)
    \item $\mathcal{D}_{\mathrm{tar},3}$: balanced mixture of three tasks ($\bm{a}_{\text{mix}} = (1/3, 1/3, 1/3)$)
    \item $\mathcal{D}_{\mathrm{tar},4}$: moderately imbalanced mixture of three tasks ($\bm{a}_{\text{mix}} = (0.4, 0.4, 0.2)$)
    \item $\mathcal{D}_{\mathrm{tar},5}$: highly imbalanced mixture of three tasks ($\bm{a}_{\text{mix}} = (0.6, 0.2, 0.2)$)
\end{enumerate}
We randomly down-sample the test datasets to make their size identical among the five configurations. The total numbers of test data are 1,000 for CIFAR-100-5 and ImageNet-R-5, 500 for CIFAR-100-20 and ImageNet-R-20, and 200 for CIFAR-100-50 and ImageNet-R-50.
For ImageNet-R-20 and ImageNet-R-50, the mixing ratio could not be strictly satisfied due to the shortage of test data; however, it was followed as much as possible.
We split the down-sampled dataset and use 10\% of it as the meta dataset. The remaining 90\% are used for the model evaluation. 
In addition, 500 images sampled from each training task are used to calculate dataset similarity in all experiments.
For each configuration, we generate five variants by varying random seeds for the selection of $M$ tasks and data sampling, and use the average top-1 accuracy across these five variants as a performance measure.
We performed three trials for each setting by changing the random seeds for the training and model merge processes.

\input{table_cifar100-20}
We compare Tunable MAGMAX against \textbf{Baseline}, which sequentially fine-tunes the model on each task without any mitigation of CF, as well as existing model merging approaches: \textbf{Rand Mix}~\cite{marczak_magmax_2025}, Model Soup (\textbf{Average})~\cite{modelsoup}, \textbf{TIES Merging}~\cite{ties_merging}, and \textbf{MAGMAX}~\cite{marczak_magmax_2025}. 
We set the scaling factor as $\lambda_{\mathrm{merge}}=0.5$ for all model merging methods. 
For our proposed method, we prepare two variations for computing the preference vector introduced in Section~\ref{sec:proposed:main}: OT distance-based preference control (\textbf{OT}) and label distribution-based preference control (\textbf{Label}). 

Tables~\ref{tab:cifar100-20} and~\ref{tab:imagenetr-5} present the average accuracy on different target environments in CIFAR-100-20 and ImageNet-R-5, respectively. The results on the other settings are provided in supplementary materials. 
Tunable MAGMAX (Label) and Tunable MAGMAX (OT) consistently achieve higher or comparable performance across all experimental configurations compared to the other methods.

\input{table_imagenetr-5}

\subsection{Ablation Study}\label{sec:ablation}

\subsubsection{Sensitivity for Dataset Similarity Methods}\label{sec:exp:similarity}

We compare different methods to measure the similarity between the target environment and the training tasks.
Specifically, we evaluate optimal transport distance (\textbf{OT}), cosine distance (\textbf{Cos}), and maximum mean discrepancy (\textbf{MMD}) with the RBF kernel as dataset distance metrics.
For the cosine distance, we compute the mean embedding vector from the feature representations of each dataset and calculate the cosine distance between the two mean vectors.

\begin{wraptable}{r}{0.45\linewidth}
\centering
\vspace{-20pt}
\input{table_similarity}

\label{tab:similarity}
\vspace{-10pt}
\end{wraptable}

Table~\ref{tab:similarity} shows the average accuracy in all target
environments $\left\{ \mathcal{D}_{\mathrm{tar},i} \right\}_{i=1}^5$ for CIFAR-100-$T$.
It demonstrates that Tunable MAGMAX (Label) achieves superior performance to the others across different settings, as it  accurately reflects the target data distribution and consequently produces merged models with the highest performance for the target environment.
Among the label-free methods, Tunable MAGMAX (OT) consistently outperforms the others, suggesting that the OT distance effectively captures the similarity between datasets for the construction of the preference vector.

\subsubsection{Robustness for Diversity of Target Environments}
\begin{wrapfigure}{r}{0.41\linewidth}
    \centering
    \vspace{-23pt}
    \includegraphics[width=\linewidth]{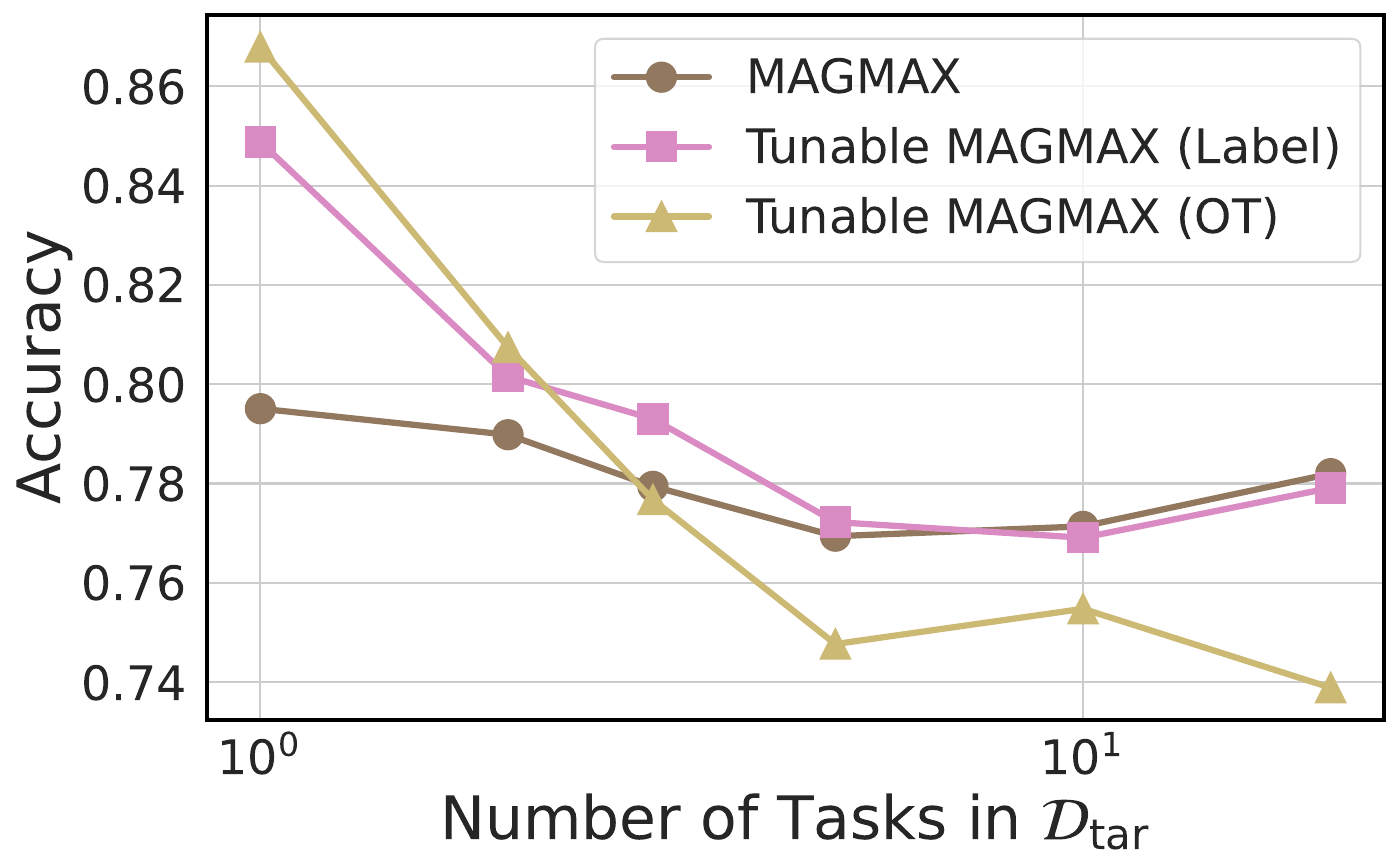}
    \caption{Average accuracy with different numbers of tasks in target environments in CIFAR-100-20}
    \label{fig:num_tasks_in_target}
    \vspace{-20pt}
\end{wrapfigure}

Figure~\ref{fig:num_tasks_in_target} illustrates how performance changes as the number of tasks included in the target environment varies for CIFAR-100-20.
Within the target environment data, the number of samples for each task is kept equal, that is, we set $\bm{a}_\text{mix} = (1/M, \ldots, 1/M)$ for $M \in \{1,2,3,5,10,20\}$.
We report the average accuracy over three variations of task selection for each value of $M$.

The results show that Tunable MAGMAX (Label) consistently achieves high performance across varying numbers of tasks in the target environment, especially when the target environment involves a small number of tasks. 
Tunable MAGMAX (OT) also shows high accuracy when $M$ is small, while its performance deteriorates when $M$ becomes large.
These results highlight the robustness of the label-based preference control compared to the OT distance-based one.

\section{Conclusion}
In this work, we proposed a preference-aware framework called Tunable MAGMAX for constructing continual learning models that adapt to various target environments.
We introduced a novel model merging method that controls the number of elements selected from each task vector according to the preference vector, enabling flexible adjustment of task-specific performance.
We further extend our framework to automatically construct an appropriate preference vector using the dataset similarities between the meta dataset collected from the target environment and datasets from the model training tasks.
The experimental results demonstrated that our proposed method constructs suitable merged models according to the preference vector and achieves superior performance compared to other methods in various target environments.

The flexibility of our approach allows practitioners with limited computational resources to build suitable merged models according to their specific requirements, making it a practical solution for continual learning scenarios where deployment needs vary across applications and users.

We conclude by discussing some limitations and future directions of our approach. First, to construct a preference vector based on the dataset similarity, our approach requires a certain amount of meta- and training data collected from the target environment and the model training tasks enough to approximate the whole environment or tasks. Second, our construction mechanism for the preference vector does not consider dependencies among training tasks, even though there exist cases where some fine-tuned models in later tasks may partially preserve the performance on earlier tasks.
Some potential future works include modifying the mechanism of preference vector construction by considering the task order and the potential of each task vector for other tasks.

\begin{credits}
\subsubsection{\ackname}
This study was partially funded by JSPS KAKENHI (JP23K28156 and JP23H00491).
\end{credits}

\bibliographystyle{splncs04}
\bibliography{mybibliography}

\appendix

\section{Results in Other Dataset Settings} \label{app:main_results_other} 

We report our main results with the experimental settings in Section 4.3 for CIFAR-100-5, CIFAR-100-50, ImageNet-R-20, and ImageNet-R-50. 
For ImageNet-R-20 and ImageNet-R-50, our proposed methods perform comparably to existing methods. 
This is primarily due to the limited training data per task in these settings.
With these settings, each task vector exhibits reduced representation of the corresponding task, resulting in diminishing our method's strength.
The competitive performance of baseline methods further supports this interpretation.

\input{table_cifar100-5}
\input{table_cifar100-50}
\input{table_imagenetr-20}
\input{table_imagenetr-50}

\end{document}

%% file: macro.tex
\usepackage{stmaryrd}
\usepackage{amsmath}
\usepackage{amsfonts}

\SetSymbolFont{stmry}{bold}{U}{stmry}{m}{n}

\newcommand{\argmax}{\mathop{\rm argmax}\limits}

\usepackage{xcolor}
\usepackage{ifthen}
\usepackage{comment}
\usepackage{amsmath, amssymb}
\usepackage{type1cm}
\usepackage{algorithmic}
\usepackage{algorithm}
\usepackage{bm}
\usepackage{bbm}
\usepackage{booktabs}
\usepackage{multirow}
\usepackage{url}
\usepackage{appendix}
\usepackage{subcaption}

\newcommand{\markupdraft}[2]{
    \ifthenelse{\equal{#1}{display}}{#2}{}
    \ifthenelse{\equal{#1}{color}}{\color{#2}}{}
}

\newcommand{\newcolored}[3][]{{\markupdraft{color}{#2}#3}
    \ifthenelse{\equal{#1}{}}{}{\markupdraft{display}{{\color{yellow!70!black}[#1]}}}}
\newcommand{\del}[2][]{{\markupdraft{display}{{\color{orange}[removed: ``#2''[#1]]}}}} 

\renewcommand{\del}[2]{}  

%% file: pseudo_whole.tex
\begin{algorithm}[t!]
	\caption{Tunable MAGMAX with Similarity-based Preference Control}
	\label{alg:whole}
    \begin{algorithmic}[1]
    \REQUIRE Weight parameter of LPM $\bm{\theta}_0 \in \mathbb{R}^d$, task sequence $\{\mathcal{D}_t\}_{t=1}^T$, small amount of data $\mathcal{D}_{\mathrm{meta}}$ from deployment environments, merge coefficient $\lambda_{\mathrm{merge}} \in [0,1]$
    
    \FOR{$t = 1,\ldots, T$}
        \STATE Get $\bm{\theta}_t$ by fine-tuning with $\mathcal{D}_t$ from $\bm{\theta}_{t-1}$.
        \STATE Compute task vector as $\bm{\tau}_t = \bm{\theta}_t-\bm{\theta}_0$.
        \STATE Compute dataset similarity $s_t = \mathrm{sim}(\mathcal{D}_t,\mathcal{D}_{\mathrm{meta}})$ as described in Section~\ref{sec:proposed:main}.
    \ENDFOR
    
    \STATE $\bm{n}_T \leftarrow$ Compute preference vector using similarities $\{ s_t \}^T_{t=1}$ with \eqref{eq:sim_to_pv} and \eqref{eq:remainder}.
    \STATE $\bm{\tau}_\mathrm{merged} \leftarrow$ Merge task vectors $\{\bm{\tau}_t\}_{t=1}^T$ using Tunable MAGMAX with $\bm{n}_T$.
    \STATE Compute $\bm{\theta}_\mathrm{merged} = \bm{\theta}_0 + \lambda_{\mathrm{merge}} \bm{\tau}_\mathrm{merged}$ 
    \RETURN $\bm{\theta}_\mathrm{merged}$
\end{algorithmic}
\end{algorithm}

%% file: table_cifar100-20.tex
\begin{table}[t!]
\centering
\caption{Average accuracy on the target datasets in CIFAR-100-20}
\label{tab:cifar100-20}
\vspace{1em}
\small
\begin{tabular}{l @{\hspace{3pt}}|@{\hspace{3pt}} l ccccc c}
\toprule
\multicolumn{2}{c}{} & $\mathcal{D}_{\mathrm{tar},1}$ & $\mathcal{D}_{\mathrm{tar},2}$ & $\mathcal{D}_{\mathrm{tar},3}$ & $\mathcal{D}_{\mathrm{tar},4}$ & $\mathcal{D}_{\mathrm{tar},5}$ & \multirow{2}{*}{Average} \\
\multicolumn{2}{l}{{\scriptsize Mixing Ratio $\bm{a}_\text{mix}$}} & {\scriptsize $(0.5,0.5)$} & {\scriptsize $(0.8,0.2)$} & {\scriptsize $(1/3,1/3,1/3)$} & {\scriptsize $(0.4,0.4,0.2)$} & {\scriptsize $(0.6,0.2,0.2)$} & \\
\midrule
\multicolumn{2}{l}{Baseline} & 0.78 {\scriptsize $\pm$0.01} & 0.78 {\scriptsize $\pm$0.00} & 0.76 {\scriptsize ± 0.01} & 0.76 {\scriptsize $\pm$0.00} & 0.77 {\scriptsize $\pm$0.01} & 0.77 \\
\multicolumn{2}{l}{Random Mix} & 0.77 {\scriptsize $\pm$0.00} & 0.77 {\scriptsize $\pm$0.02} & 0.75 {\scriptsize $\pm$0.02} & 0.77 {\scriptsize $\pm$0.01} & 0.76 {\scriptsize $\pm$0.02} & 0.76 \\
\multicolumn{2}{l}{Average} & 0.77 {\scriptsize $\pm$0.01} & 0.77 {\scriptsize $\pm$0.02} & 0.75 {\scriptsize $\pm$0.01} & 0.77 {\scriptsize $\pm$0.01} & 0.76 {\scriptsize $\pm$0.02} & 0.76 \\
\multicolumn{2}{l}{TIES-Merging} & 0.77 {\scriptsize $\pm$0.01} & 0.78 {\scriptsize $\pm$0.02} & 0.76 {\scriptsize $\pm$0.01} & 0.77 {\scriptsize $\pm$0.00} & 0.76 {\scriptsize $\pm$0.02} & 0.77 \\
\multicolumn{2}{l}{MAGMAX} & 0.78 {\scriptsize $\pm$0.01} & 0.78 {\scriptsize $\pm$0.02} & 0.77 {\scriptsize $\pm$0.02} & 0.78 {\scriptsize $\pm$0.01} & 0.76 {\scriptsize $\pm$0.02} & 0.77 \\
\midrule
\multirow{2}{*}{\shortstack[l]{Tunable\\ MAGMAX}} & OT & 0.79 {\scriptsize $\pm$0.02} & 0.81 {\scriptsize $\pm$0.01} & 0.78 {\scriptsize $\pm$0.01} & 0.80 {\scriptsize $\pm$0.02} & 0.78 {\scriptsize $\pm$0.02} & 0.79 \\
& Label & \textbf{0.81} {\scriptsize $\pm$0.03} & \textbf{0.82} {\scriptsize $\pm$0.02} & \textbf{0.80} {\scriptsize $\pm$0.03} & \textbf{0.82} {\scriptsize $\pm$0.02} & \textbf{0.81} {\scriptsize $\pm$0.01} & \textbf{0.81} \\
\bottomrule
\end{tabular}
\end{table}

%% file: table_imagenetr-5.tex
\begin{table}[t!]
\centering
\caption{Average accuracy on the target datasets in ImageNet-R-5}
\label{tab:imagenetr-5}
\vspace{1em}
\small
\begin{tabular}{l @{\hspace{3pt}}|@{\hspace{3pt}} l ccccc c}
\toprule
\multicolumn{2}{c}{} & $\mathcal{D}_{\mathrm{tar},1}$ & $\mathcal{D}_{\mathrm{tar},2}$ & $\mathcal{D}_{\mathrm{tar},3}$ & $\mathcal{D}_{\mathrm{tar},4}$ & $\mathcal{D}_{\mathrm{tar},5}$ & \multirow{2}{*}{Average} \\
\multicolumn{2}{l}{{\scriptsize Mixing Ratio $\bm{a}_\text{mix}$}} & {\scriptsize $(0.5,0.5)$} & {\scriptsize $(0.8,0.2)$} & {\scriptsize $(1/3,1/3,1/3)$} & {\scriptsize $(0.4,0.4,0.2)$} & {\scriptsize $(0.6,0.2,0.2)$} & \\
\midrule
\multicolumn{2}{l}{Baseline} & 0.84 {\scriptsize $\pm$0.00} & 0.83 {\scriptsize $\pm$0.01} & 0.84 {\scriptsize $\pm$0.01} & 0.84 {\scriptsize $\pm$0.01} & 0.84 {\scriptsize $\pm$0.00} & 0.84 \\
\multicolumn{2}{l}{Random Mix} & 0.83 {\scriptsize $\pm$0.01} & 0.83 {\scriptsize $\pm$0.01} & 0.84 {\scriptsize $\pm$0.01} & 0.83 {\scriptsize $\pm$0.00} & 0.84 {\scriptsize $\pm$0.01} & 0.83 \\
\multicolumn{2}{l}{Average} & 0.84 {\scriptsize $\pm$0.00} & 0.84 {\scriptsize $\pm$0.00} & 0.84 {\scriptsize $\pm$0.00} & 0.83 {\scriptsize $\pm$0.01} & 0.83 {\scriptsize $\pm$0.01} & 0.84 \\
\multicolumn{2}{l}{TIES-Merging} & 0.84 {\scriptsize $\pm$0.00} & 0.84 {\scriptsize $\pm$0.01} & 0.84 {\scriptsize $\pm$0.01} & 0.84 {\scriptsize $\pm$0.00} & 0.84 {\scriptsize $\pm$0.01} & 0.84 \\
\multicolumn{2}{l}{MAGMAX} & \textbf{0.85} {\scriptsize $\pm$0.00} & 0.84 {\scriptsize $\pm$0.01} & \textbf{0.85} {\scriptsize $\pm$0.01} & 0.84 {\scriptsize $\pm$0.01} & \textbf{0.85} {\scriptsize $\pm$0.01} & \textbf{0.85} \\
\midrule
\multirow{2}{*}{\shortstack[l]{Tunable\\ MAGMAX}} & OT & 0.84 {\scriptsize $\pm$0.00} & \textbf{0.85} {\scriptsize $\pm$0.00} & 0.83 {\scriptsize $\pm$0.01} & \textbf{0.85} {\scriptsize $\pm$0.00} & 0.83 {\scriptsize $\pm$0.01} & 0.84 \\
& Label & \textbf{0.85} {\scriptsize $\pm$0.00} & \textbf{0.85} {\scriptsize $\pm$0.01} & 0.84 {\scriptsize $\pm$0.01} & \textbf{0.85} {\scriptsize $\pm$0.01} & \textbf{0.85} {\scriptsize $\pm$0.01} & \textbf{0.85} \\
\bottomrule
\end{tabular}
\end{table}

%% file: table_similarity.tex
\normalsize
\caption{Average accuracy on all target environments in CIFAR-100-$T$}

\small 
\begin{tabular}{c | c cccc}
\hline
\multirow{2}{*}{$T$} & \multirow{2}{*}{MAGMAX} & \multicolumn{4}{c}{Tunable MAGMAX} \\
& & Cos & MMD & OT & Label  \\
\hline
5 & 0.83 & 0.84 & 0.80 & 0.83 & \textbf{0.85}  \\
20 & 0.77 & 0.78 & 0.74 & 0.79 & \textbf{0.81} \\
50 & 0.77 & 0.75 & 0.74 & 0.76 & \textbf{0.78} \\
\hline
\end{tabular}

%% file: table_cifar100-5.tex
\begin{table}[h!]
\centering
\caption{Average accuracy on the target datasets in CIFAR-100-5}
\vspace{1em}
\small
\begin{tabular}{l @{\hspace{3pt}}|@{\hspace{3pt}} l ccccc c}
\toprule
\multicolumn{2}{c}{} & $\mathcal{D}_{\mathrm{tar},1}$ & $\mathcal{D}_{\mathrm{tar},2}$ & $\mathcal{D}_{\mathrm{tar},3}$ & $\mathcal{D}_{\mathrm{tar},4}$ & $\mathcal{D}_{\mathrm{tar},5}$ & \multirow{2}{*}{Average} \\
\multicolumn{2}{l}{{\scriptsize Mixing Ratio $\bm{a}_\text{mix}$}} & {\scriptsize $(0.5,0.5)$} & {\scriptsize $(0.8,0.2)$} & {\scriptsize $(1/3,1/3,1/3)$} & {\scriptsize $(0.4,0.4,0.2)$} & {\scriptsize $(0.6,0.2,0.2)$} & \\
\midrule
\multicolumn{2}{l}{Baseline} & 0.82 {\scriptsize $\pm$0.01} & 0.82 {\scriptsize $\pm$0.01} & 0.81 {\scriptsize $\pm$0.01} & 0.83 {\scriptsize $\pm$0.01} & 0.82 {\scriptsize $\pm$0.01} & 0.82 \\
\multicolumn{2}{l}{Random Mix} & 0.82 {\scriptsize $\pm$0.01} & 0.82 {\scriptsize $\pm$0.01} & 0.81 {\scriptsize $\pm$0.01} & 0.80 {\scriptsize $\pm$0.01} & 0.81 {\scriptsize $\pm$0.00} & 0.81 \\
\multicolumn{2}{l}{Average} & 0.82 {\scriptsize $\pm$0.01} & 0.81 {\scriptsize $\pm$0.01} & 0.81 {\scriptsize $\pm$0.01} & 0.80 {\scriptsize $\pm$0.01} & 0.81 {\scriptsize $\pm$0.00} & 0.81 \\
\multicolumn{2}{l}{TIES-Merging} & 0.83 {\scriptsize $\pm$0.01} & 0.82 {\scriptsize $\pm$0.01} & 0.82 {\scriptsize $\pm$0.01} & 0.81 {\scriptsize $\pm$0.01} & 0.81 {\scriptsize $\pm$0.00} & 0.82 \\
\multicolumn{2}{l}{MAGMAX} & 0.83 {\scriptsize $\pm$0.01} & 0.83 {\scriptsize $\pm$0.00} & 0.83 {\scriptsize $\pm$0.01} & 0.83 {\scriptsize $\pm$0.01} & 0.83 {\scriptsize $\pm$0.00} & 0.83 \\
\midrule
\multirow{2}{*}{\shortstack[l]{Tunable\\ MAGMAX}} & OT & 0.81 {\scriptsize $\pm$0.01} & \textbf{0.87} {\scriptsize $\pm$0.00} & 0.81 {\scriptsize $\pm$0.01} & 0.83 {\scriptsize $\pm$0.01} & 0.80 {\scriptsize $\pm$0.00} & 0.83 \\
& Label & \textbf{0.85} {\scriptsize $\pm$0.01} & 0.86 {\scriptsize $\pm$0.01} & \textbf{0.83} {\scriptsize $\pm$0.00} & \textbf{0.85} {\scriptsize $\pm$0.00} & \textbf{0.85} {\scriptsize $\pm$0.01} & \textbf{0.85} \\
\bottomrule
\end{tabular}
\end{table}

%% file: table_cifar100-50.tex
\begin{table}[h!]
\centering
\caption{Average accuracy on the target datasets in CIFAR-100-50}
\vspace{1em}
\small
\begin{tabular}{l @{\hspace{3pt}}|@{\hspace{3pt}} l ccccc c}
\toprule
\multicolumn{2}{c}{} & $\mathcal{D}_{\mathrm{tar},1}$ & $\mathcal{D}_{\mathrm{tar},2}$ & $\mathcal{D}_{\mathrm{tar},3}$ & $\mathcal{D}_{\mathrm{tar},4}$ & $\mathcal{D}_{\mathrm{tar},5}$ & \multirow{2}{*}{Average} \\
\multicolumn{2}{l}{{\scriptsize Mixing Ratio $\bm{a}_\text{mix}$}} & {\scriptsize $(0.5,0.5)$} & {\scriptsize $(0.8,0.2)$} & {\scriptsize $(1/3,1/3,1/3)$} & {\scriptsize $(0.4,0.4,0.2)$} & {\scriptsize $(0.6,0.2,0.2)$} & \\
\midrule
\multicolumn{2}{l}{Baseline} & 0.77 {\scriptsize $\pm$0.01} & 0.74 {\scriptsize $\pm$0.02} & 0.74 {\scriptsize $\pm$0.03} & 0.75 {\scriptsize $\pm$0.04} & 0.78 {\scriptsize $\pm$0.04} & 0.76 \\
\multicolumn{2}{l}{Random Mix} & 0.74 {\scriptsize $\pm$0.03} & 0.75 {\scriptsize $\pm$0.02} & 0.73 {\scriptsize $\pm$0.02} & 0.73 {\scriptsize $\pm$0.01} & 0.76 {\scriptsize $\pm$0.04} & 0.74 \\
\multicolumn{2}{l}{Average} & 0.75 {\scriptsize $\pm$0.03} & 0.74 {\scriptsize $\pm$0.01} & 0.73 {\scriptsize $\pm$0.03} & 0.73 {\scriptsize $\pm$0.02} & 0.77 {\scriptsize $\pm$0.03} & 0.74 \\
\multicolumn{2}{l}{TIES-Merging} & 0.77 {\scriptsize $\pm$0.03} & 0.76 {\scriptsize $\pm$0.02} & 0.72 {\scriptsize $\pm$0.02} & \textbf{0.79} {\scriptsize $\pm$0.05} & 0.77 {\scriptsize $\pm$0.00} & 0.76 \\
\multicolumn{2}{l}{MAGMAX} & 0.77 {\scriptsize $\pm$0.01} & 0.77 {\scriptsize $\pm$0.01} & 0.75 {\scriptsize $\pm$0.02} & 0.75 {\scriptsize $\pm$0.02} & 0.78 {\scriptsize $\pm$0.02} & 0.77 \\
\midrule
\multirow{2}{*}{\shortstack[l]{Tunable\\ MAGMAX}} & OT & 0.76 {\scriptsize $\pm$0.01} & 0.75 {\scriptsize $\pm$0.02} & 0.74 {\scriptsize $\pm$0.05} & 0.78 {\scriptsize $\pm$0.04} & 0.76 {\scriptsize $\pm$0.07} & 0.76 \\
& Label & \textbf{0.78} {\scriptsize $\pm$0.03} & \textbf{0.80} {\scriptsize $\pm$0.04} & \textbf{0.78} {\scriptsize $\pm$0.04} & 0.77 {\scriptsize $\pm$0.06} & \textbf{0.79} {\scriptsize $\pm$0.02} & \textbf{0.78} \\
\bottomrule
\end{tabular}
\end{table}

%% file: table_imagenetr-20.tex
\begin{table}[!ht]
\centering
\caption{Average accuracy on the target datasets in ImageNet-R-20}
\vspace{1em}
\small
\begin{tabular}{l @{\hspace{3pt}}|@{\hspace{3pt}} l ccccc c}
\toprule
\multicolumn{2}{c}{} & $\mathcal{D}_{\mathrm{tar},1}$ & $\mathcal{D}_{\mathrm{tar},2}$ & $\mathcal{D}_{\mathrm{tar},3}$ & $\mathcal{D}_{\mathrm{tar},4}$ & $\mathcal{D}_{\mathrm{tar},5}$ & \multirow{2}{*}{Average} \\
\multicolumn{2}{l}{{\scriptsize Mixing Ratio $\bm{a}_\text{mix}$}} & {\scriptsize $(0.5,0.5)$} & {\scriptsize $(0.8,0.2)$} & {\scriptsize $(1/3,1/3,1/3)$} & {\scriptsize $(0.4,0.4,0.2)$} & {\scriptsize $(0.6,0.2,0.2)$} & \\
\midrule
\multicolumn{2}{l}{Baseline} & 0.83 {\scriptsize $\pm$0.02} & 0.81 {\scriptsize $\pm$0.01} & 0.82 {\scriptsize $\pm$0.01} & 0.82 {\scriptsize $\pm$0.01} & 0.81 {\scriptsize $\pm$0.02} & 0.82 \\
\multicolumn{2}{l}{Random Mix} & 0.82 {\scriptsize $\pm$0.01} & 0.81 {\scriptsize $\pm$0.01} & 0.81 {\scriptsize $\pm$0.01} & 0.82 {\scriptsize $\pm$0.00} & 0.81 {\scriptsize $\pm$0.02} & 0.81 \\
\multicolumn{2}{l}{Average} & 0.82 {\scriptsize $\pm$0.01} & 0.81 {\scriptsize $\pm$0.01} & 0.81 {\scriptsize $\pm$0.01} & 0.82 {\scriptsize $\pm$0.01} & 0.81 {\scriptsize $\pm$0.01} & 0.81 \\
\multicolumn{2}{l}{TIES-Merging} & 0.83 {\scriptsize $\pm$0.02} & \textbf{0.82} {\scriptsize $\pm$0.00} & 0.82 {\scriptsize $\pm$0.01} & 0.82 {\scriptsize $\pm$0.01} & 0.81 {\scriptsize $\pm$0.01} & 0.82 \\
\multicolumn{2}{l}{MAGMAX} & \textbf{0.84} {\scriptsize $\pm$0.01} & \textbf{0.82} {\scriptsize $\pm$0.00} & \textbf{0.83} {\scriptsize $\pm$0.01} & \textbf{0.83} {\scriptsize $\pm$0.01} & 0.82 {\scriptsize $\pm$0.02} & \textbf{0.83} \\
\midrule
\multirow{2}{*}{\shortstack[l]{Tunable\\ MAGMAX}} & OT & 0.83 {\scriptsize $\pm$0.01} & \textbf{0.82} {\scriptsize $\pm$0.01} & \textbf{0.83} {\scriptsize $\pm$0.01} & 0.82 {\scriptsize $\pm$0.00} & 0.82 {\scriptsize $\pm$0.01} & 0.82 \\
& Label & 0.83 {\scriptsize $\pm$0.01} & \textbf{0.82} {\scriptsize $\pm$0.01} & 0.82 {\scriptsize $\pm$0.01} & 0.82 {\scriptsize $\pm$0.02} & \textbf{0.83} {\scriptsize $\pm$0.02} & \textbf{0.83} \\
\bottomrule
\end{tabular}
\end{table}

%% file: table_imagenetr-50.tex
\begin{table}[!ht]
\centering
\caption{Average accuracy on thet target datasets in ImageNet-R-50}
\vspace{1em}
\small
\begin{tabular}{l @{\hspace{3pt}}|@{\hspace{3pt}} l ccccc c}
\toprule
\multicolumn{2}{c}{} & $\mathcal{D}_{\mathrm{tar},1}$ & $\mathcal{D}_{\mathrm{tar},2}$ & $\mathcal{D}_{\mathrm{tar},3}$ & $\mathcal{D}_{\mathrm{tar},4}$ & $\mathcal{D}_{\mathrm{tar},5}$ & \multirow{2}{*}{Average} \\
\multicolumn{2}{l}{{\scriptsize Mixing Ratio $\bm{a}_\text{mix}$}} & {\scriptsize $(0.5,0.5)$} & {\scriptsize $(0.8,0.2)$} & {\scriptsize $(1/3,1/3,1/3)$} & {\scriptsize $(0.4,0.4,0.2)$} & {\scriptsize $(0.6,0.2,0.2)$} & \\
\midrule
\multicolumn{2}{l}{Baseline} & 0.82 {\scriptsize $\pm$0.04} & 0.82 {\scriptsize $\pm$0.02} & 0.80 {\scriptsize $\pm$0.01} & 0.82 {\scriptsize $\pm$0.02} & 0.81 {\scriptsize $\pm$0.01} & \textbf{0.82} \\
\multicolumn{2}{l}{Random Mix} & \textbf{0.83} {\scriptsize $\pm$0.04} & 0.82 {\scriptsize $\pm$0.02} & 0.79 {\scriptsize $\pm$0.01} & 0.82 {\scriptsize $\pm$0.01} & 0.80 {\scriptsize $\pm$0.01} & 0.81 \\
\multicolumn{2}{l}{Average} & 0.82 {\scriptsize $\pm$0.04} & 0.81 {\scriptsize $\pm$0.02} & 0.79 {\scriptsize $\pm$0.01} & 0.82 {\scriptsize $\pm$0.02} & 0.80 {\scriptsize $\pm$0.00} & 0.81 \\
\multicolumn{2}{l}{TIES-Merging} & 0.82 {\scriptsize $\pm$0.04} & 0.81 {\scriptsize $\pm$0.03} & 0.79 {\scriptsize $\pm$0.01} & 0.82 {\scriptsize $\pm$0.03} & 0.80 {\scriptsize $\pm$0.01} & 0.81 \\
\multicolumn{2}{l}{MAGMAX} & \textbf{0.83} {\scriptsize $\pm$0.03} & \textbf{0.83} {\scriptsize $\pm$0.02} & \textbf{0.81} {\scriptsize $\pm$0.01} & \textbf{0.83} {\scriptsize $\pm$0.02} & \textbf{0.82} {\scriptsize $\pm$0.01} & \textbf{0.82} \\
\midrule
\multirow{2}{*}{\shortstack[l]{Tunable\\ MAGMAX}} & OT & 0.80 {\scriptsize $\pm$0.02} & 0.79 {\scriptsize $\pm$0.03} & 0.80 {\scriptsize $\pm$0.03} & 0.80 {\scriptsize $\pm$0.02} & 0.80 {\scriptsize $\pm$0.01} & 0.80 \\
& Label & 0.82 {\scriptsize $\pm$0.02} & 0.79 {\scriptsize $\pm$0.04} & \textbf{0.81} {\scriptsize $\pm$0.00} & 0.80 {\scriptsize $\pm$0.01} & 0.81 {\scriptsize $\pm$0.02} & 0.81 \\
\bottomrule
\end{tabular}
\end{table}